\documentclass[letterpaper, 10 pt, conference]{ieeeconf}  % Comment this line out if you need a4paper

\IEEEoverridecommandlockouts                              % This command is only needed if 
\usepackage{cite}
\usepackage{amsmath,amssymb,amsfonts}
\usepackage{graphicx}
\usepackage{textcomp}
\usepackage{xcolor}

\usepackage{makeidx}

\usepackage{pgfplots}
\pgfplotsset{compat=newest}
\usepackage{multirow}
\usepackage{tabularx}
\usepackage{threeparttable}
\usepackage{booktabs}

\usepackage[draft,inline,nomargin]{fixme}
\fxusetheme{color}

\usepackage{array}
\usepackage{float}
\usepackage{graphbox}
\usepackage{subcaption}
\usepackage{tikz,tikz-3dplot}
\usetikzlibrary{fit,arrows,arrows.meta,automata,backgrounds,calc,chains,%
decorations.markings,decorations.pathreplacing,decorations.pathmorphing,%
matrix,positioning,shapes,shapes.geometric,shapes.symbols,spy,trees,tikzmark}
\usepackage{balance}
\usepackage[binary-units=true,product-units=single,per-mode=symbol,range-units=single,range-phrase=\,--\,,detect-all]{siunitx}
\DeclareSIUnit\pixel{px}
\usepackage{bm}
\usepackage{acronym}
\usepackage{tablefootnote}
\usepackage{hyperref}

\usepackage{caption}
\let\labelindent\relax
\usepackage[inline]{enumitem}

\usepackage{adjustbox}
\newcolumntype{R}[2]{%
    >{\adjustbox{angle=#1,lap=\width-(#2)}\bgroup}%
    l%
    <{\egroup}%
}
\newcolumntype{L}[1]{>{\raggedright\let\newline\\\arraybackslash\hspace{0pt}}m{#1}}

\title{\LARGE \bf
SWA-SOP: Spatially-aware Window Attention for Semantic Occupancy Prediction in Autonomous Driving
}

\author{Helin Cao$^{*1,2,3}$, Rafael Materla$^{*1}$, and Sven Behnke$^{1,2,3}$% <-this % stops a space
\thanks{*Equal contribution.}% <-this % stops a space
\thanks{$^{1}$Autonomous Intelligent Systems group, Computer Science Institute VI – Intelligent Systems and Robotics, University of Bonn, Germany {\tt\small cao@ais.uni-bonn.de}}%
\thanks{$^{2}$Center for Robotics, University of Bonn, Germany}%
\thanks{$^{3}$Lamarr Institute for Machine Learning and Artificial Intelligence, Germany}%
}

\begin{document}

\maketitle
\thispagestyle{empty}
\pagestyle{empty}

%%%%%%%%%%%%%%%%%%%%%%%%%%%%%%%%%%%%%%%%%%%%%%%%%%%%%%%%%%%%%%%%%%%%%%%%%%%%%%%%
\begin{abstract}
Perception systems in autonomous driving rely on sensors such as LiDAR and cameras to perceive the 3D environment. However, due to occlusions and data sparsity, these sensors often fail to capture complete information. Semantic Occupancy Prediction (SOP) addresses this challenge by inferring both occupancy and semantics of unobserved regions. Existing transformer-based SOP methods lack explicit modeling of spatial structure in attention computation, resulting in limited geometric awareness and poor performance in sparse or occluded areas. To this end, we propose Spatially-aware Window Attention (SWA), a novel mechanism that incorporates local spatial context into attention. SWA significantly improves scene completion and achieves state-of-the-art results on LiDAR-based SOP benchmarks. We further validate its generality by integrating SWA into a camera-based SOP pipeline, where it also yields consistent gains across modalities.
\end{abstract}

\section{Introduction}
Autonomous vehicles rely on sensors such as LiDAR and cameras to perceive their surroundings. However, these sensors are inherently constrained: Their measurements into 3D space are sparse and occlusions from nearby objects often block critical parts of the scene. These limitations result in incomplete geometric and appearance observations, which can degrade perception reliability and hinder safe operation in complex traffic environments. To support robust decision-making and ensure driving safety, it is crucial to infer missing information and reconstruct a more complete understanding of the scene.

\begin{figure}[t]
    \centering
    \begin{subfigure}[b]{0.20\textwidth}
        \centering
        \includegraphics[width=\textwidth, keepaspectratio]{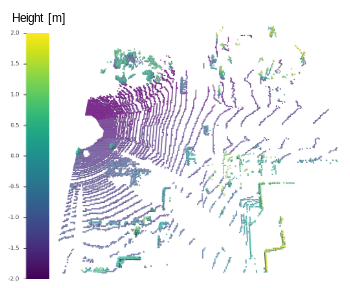}
        \caption{Sparse LiDAR input}
        \label{fig:input}
    \end{subfigure}
    \begin{subfigure}[b]{0.25\textwidth} 
        \centering
        \includegraphics[width=\textwidth, keepaspectratio]{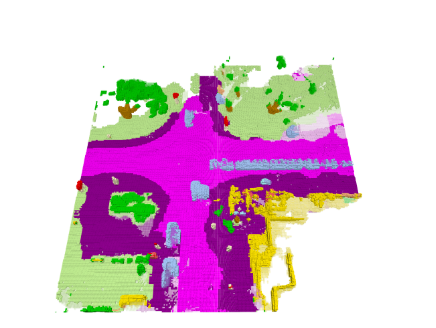}
        \caption{Dense semantic prediction}
        \label{fig:pred}
    \end{subfigure}
    
    \vspace{0.8em}
    
    \begin{subfigure}[b]{0.4\textwidth}
        \centering
        \includegraphics[width=\textwidth, keepaspectratio]{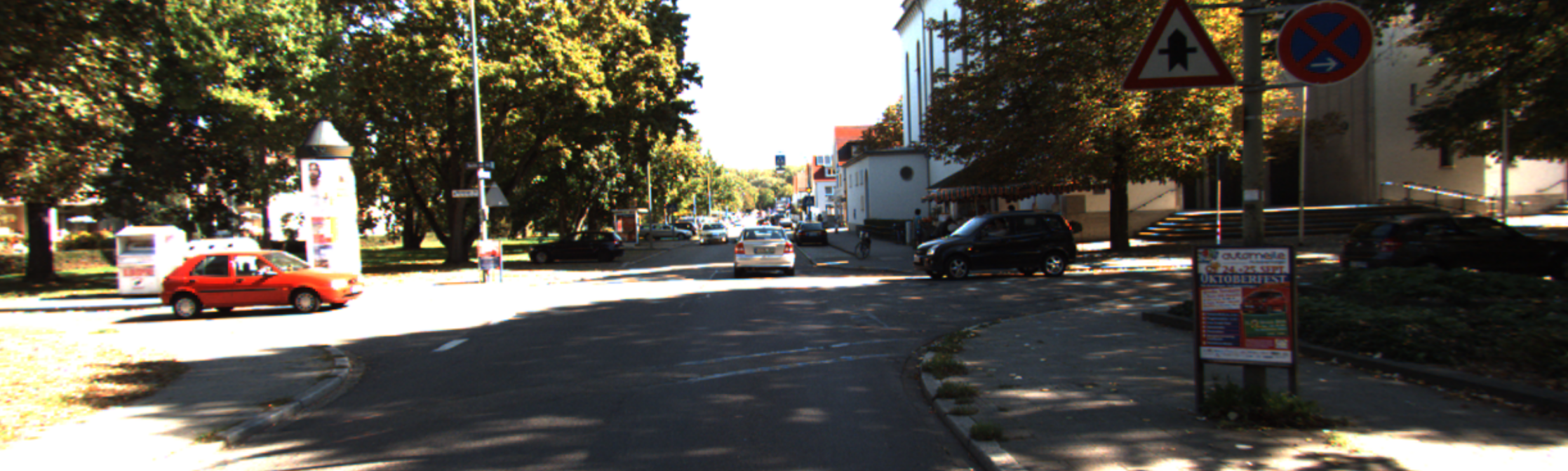}
        \caption{RGB image of a traffic scene}
        \label{fig:rgb}
    \end{subfigure}
    \caption{SWA-SOP estimates a dense, semantically labeled scene (b) from a single-frame LiDAR input (a). The proposed SWA module can also be integrated into image-based pipelines. RGB images (c) are used for visualization and extended experiments.}
    \label{fig:teaser}
    \vspace{-1.5em}
\end{figure}

Semantic Occupancy Prediction (SOP) addresses this challenge by estimating a dense 3D voxel grid with semantic labels from a single-frame input, such as an image or a LiDAR scan. The objective is to predict both occupancy and semantic categories for all voxels, including those in occluded or unobserved regions. By transforming sparse, partial sensor observations into a structured and semantically rich 3D representation, SOP enables a more comprehensive understanding of the driving scene. This dense representation supports robust perception, planning, and interaction in complex traffic environments. As illustrated in Fig.~\ref{fig:teaser}, SOP helps to bridge perceptual gaps and contributes to a more consistent and complete scene understanding.

Recent SOP methods have adopted Transformer-based architectures~\cite{zhang2024cf, zhang2023multi, wang2025medical} to improve scene completion by leveraging attention mechanisms. However, despite their ability to capture long-range dependencies, existing Transformer-based SOP models face two critical challenges. First, they lack explicit modeling of local spatial context, which is essential for maintaining geometric consistency and accurate reasoning in 3D environments. Second, the reliance on sparse depth-guided query sampling, commonly used to reduce the high computational cost of full 3D self-attention, limits their effectiveness in distant or occluded regions where LiDAR data is extremely sparse and unevenly distributed.

To address these issues, we propose Spatially-aware Window Attention Semantic Occupancy Prediction (SWA-SOP), an attention mechanism that explicitly captures local spatial context while maintaining computational efficiency. Inspired by sparse convolution, SWA-SOP applies a sliding-window attention strategy with adaptive skipping, allowing the model to selectively process regions based on query validity. Furthermore, SWA-SOP incorporates spatial information into keys, queries, and values, enabling the attention mechanism to better model geometric relationships and preserve structural continuity in scene prediction. The proposed SWA module is simple to implement and can be readily integrated into existing SOP pipelines, consistently improving prediction accuracy across different input modalities.

In summary, our main contributions are as follows: 
\begin{enumerate} 
	\item We develop a window-based attention mechanism. Compared to kernel-based convolutions, it supports flexible feature aggregation during downsampling and more effective geometric expansion during upsampling. 
	\item We introduce the Spatially-aware Window Attention (SWA) module, which incorporates local spatial context into attention computation, enhancing spatial reasoning and structural consistency in scene prediction. 
	\item We demonstrate that SWA can be used as a plug-and-play component in existing camera-based SOP pipelines, leading to consistent improvements across input modalities. 
\end{enumerate}

\section{Related Work}
\label{sec:Related Work}
\subsection{LiDAR-based and Camera-based 3D Perception}
Processing LiDAR point clouds and images into meaningful 3D representations is a core challenge in perception. For LiDAR-based semantic segmentation and object detection, common approaches include voxel-based and point-based methods. Voxel-based models~\cite{zhou2018voxelnet, zhu2021cylindrical} convert point clouds into regular grids for efficient processing with 3D CNNs or sparse convolutions. Point-based methods~\cite{qi2017pointnet, qi2017pointnet++, li2025srkd, wang2025adaptive} operate directly on irregular point sets to preserve fine-grained geometry. Hybrid methods such as CenterPoint~\cite{yin2021center} combine both paradigms to balance accuracy and efficiency. In parallel, monocular and stereo methods~\cite{shamsafar2022mobilestereonet, wang2019pseudo, you2019pseudo} estimate depth or generate pseudo-LiDAR from RGB images. Sensor fusion techniques further enhance perception by combining LiDAR accuracy with visual semantics. SLCF-Net~\cite{cao2024slcf}, for example, introduces a sequential fusion framework for semantic scene completion. Recent advances in domain-adaptive modeling~\cite{zhou2024source} and prompt-driven reasoning~\cite{li2025self} further suggest promising directions for improving robustness and transferability in multimodal perception.

\subsection{Semantic Occupancy Prediction (SOP)}
Semantic Occupancy Prediction (SOP) was first formulated by Song et al.\cite{song2017semantic} as the task of inferring a complete, semantically labeled 3D occupancy grid from a single RGB-D image. With the introduction of SemanticKITTI\cite{behley2019iccv}, SOP has expanded to large-scale outdoor LiDAR data, motivating the development of more scalable architectures. Early methods such as LMSCNet~\cite{roldao2020lmscnet} adopted hierarchical 3D convolutions for efficient voxel processing. Recently, SOP from monocular images has attracted increasing attention. onoScene~\cite{cao2022monoscene} initiated this direction via depth-to-voxel projection, while OC-SOP~\cite{cao2025ocsop} builds on it by introducing object-centric awareness for enhanced semantic and geometric reasoning. In parallel, transformer-based methods~\cite{li2023voxformer, zhang2023occformer} refine voxel features using deformable attention~\cite{zhu2021deformable}, but often suffer from sparsity and errors in depth estimation, especially in distant or occluded regions. Alternatively, DiffSSC~\cite{cao2024diffssc} explores a generative approach based on diffusion processes, offering a different modeling paradigm at the cost of higher computational complexity.

\subsection{Window-based Computation for Sparse 3D Data}
In 3D scene understanding, data is inherently sparse, as most meaningful structures lie on 2D surfaces within 3D space. Applying dense convolutions leads to redundant computation over empty or occluded regions. Sparse convolutional neural networks (SCNNs)~\cite{graham2015sparse, graham20183d, liu2015sparse} improve efficiency by restricting operations to active voxels through a sliding-window mechanism. Inspired by this, many attention-based models adopt similar windowed strategies to reduce the computational burden of global attention. Instead of fixed kernels, they perform local attention within windows, allowing more flexible, data-adaptive feature aggregation. For example, Stratified Transformer~\cite{lai2022stratified} uses non-overlapping windows with sparse global links, SphereFormer~\cite{lai2023spherical} introduces radial windows, and VoTr~\cite{mao2021voxel} applies attention to valid voxels within sliding windows. However, these models are primarily designed for detection or segmentation, focusing only on observed regions without generating dense scene completions. In contrast, semantic occupancy prediction (SOP) requires not only flexible data association, but also the ability to reason about observed geometry and complete unobserved regions. Existing attention mechanisms often lack modeling of local spatial context, limiting their ability to capture geometric structure in sparse 3D scenes. To overcome this, we introduce Spatially-aware Window Attention (SWA), which incorporates local spatial context into attention computation, enabling both improved local reasoning and more effective geometric expansion.

\section{Method}
\label{sec:Methodology}
\begin{figure*}
	\centering
	\includegraphics[width=\linewidth]{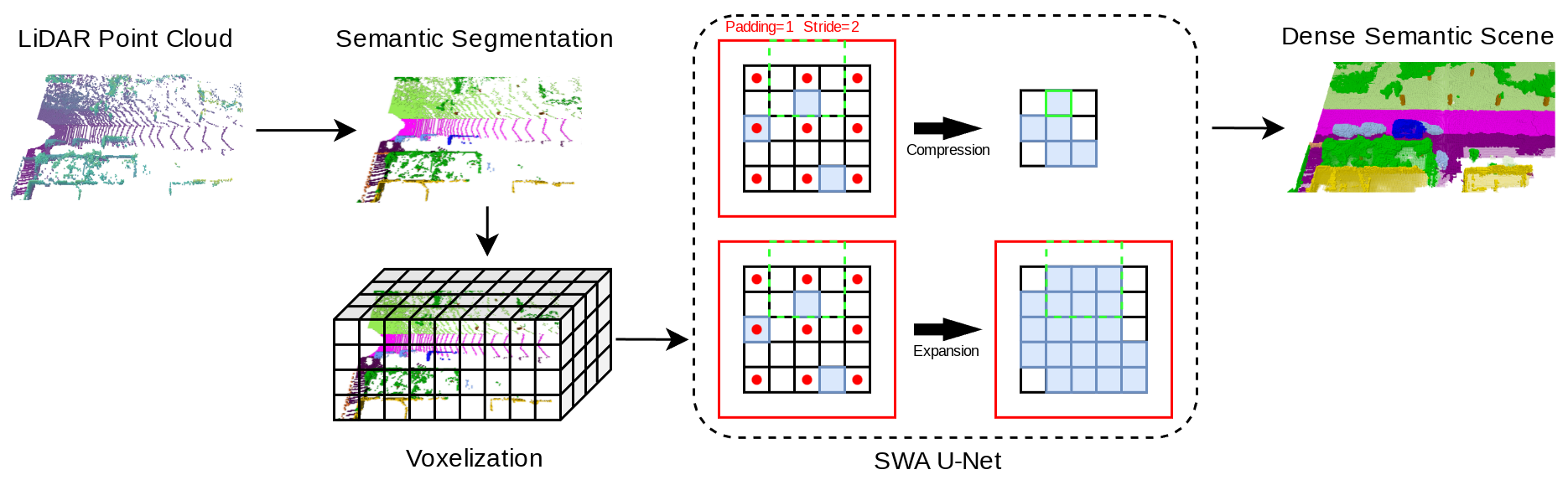}
	\caption{Overall pipeline of SWA-SOP. The raw LiDAR point cloud is first semantically segmented by SphereFormer~\cite{lai2023spherical} and then voxelized to form an initial semantic scene grid. This grid is processed by a SWA U-Net, which applies sliding-window attention inspired by sparse convolutions. During downsampling, SWA performs feature compression across valid regions; during upsampling, it supports geometric expansion by propagating features into neighboring empty voxels, enabling effective scene completion.}
	\label{fig:pipeline}
\end{figure*}

We formulate semantic occupancy prediction as the task of estimating a dense semantic volume from a sparse point cloud. Given a LiDAR point cloud $ \mathcal{X} = \{p_1, ..., p_N\} \subset \mathbb{R}^3$, where each $p_i$ denotes a 3D coordinate, the goal is to predict a semantic occupancy volume $\hat{\mathcal{Y}}$ defined over a discretized 3D space $\mathcal{V} \subset \mathbb{R}^3$. The space $\mathcal{V}$ is divided into a regular voxel grid of size $H \times W \times D$, and each voxel in $\hat{\mathcal{Y}}$ is assigned a label from the set $\mathcal{L} = \{0, 1, ..., C\}$, where $0$ indicates empty space and $1$ to $C$ correspond to semantic classes. The model aims to infer $\hat{\mathcal{Y}} \in \mathcal{L}^{H \times W \times D}
$ based on the sparse input $\mathcal{X}$, including predictions for unobserved regions, to reconstruct a complete semantic representation of the scene.

To solve this task, we design a pipeline comprised of a series of geometric and semantic reasoning stages, as illustrated in Fig.~\ref{fig:pipeline}. We first apply SphereFormer~\cite{lai2023spherical} to perform point-wise semantic segmentation on $\mathcal{X}$. To enrich the feature representation, the final semantic logits are concatenated with intermediate point-wise features extracted from SphereFormer. These enhanced point features are then voxelized to form a sparse semantic voxel grid, which serves as input to a U-Net architecture with four hierarchical levels. The U-Net processes the sparse input in a coarse-to-fine manner to generate a dense volumetric prediction.

The U-Net backbone is augmented with our proposed Spatially-aware Window Attention (SWA) module, which replaces conventional convolutions with attention-based operations. SWA is adaptable to different hierarchical levels of feature representation, making it naturally integrable into both encoder and decoder stages, unlike standard U-Nets that utilize pooling layers during up- and down-sampling. SWA performs attention-driven computations conditioned on voxel validity. Each block adopts an architecture with multi-head attention and a feedforward network, using layer normalization and residual connections in a pre-norm configuration for improved training stability. By combining sliding windows with intra-window spatial attention, SWA offers a flexible and structurally informed alternative to convolutional layers.

\subsection{Sliding Window Computation}
We define a local window $\mathcal{W}$ as a spatial subregion of size $h \times w \times d$ within the volume $\mathcal{V}$. Each window can be flattened into an ordered sequence $\mathcal{S} = [s_1, s_2, ..., s_L]$, where $L = hwd$ is the number of voxel slots. The sequence is constructed using a fixed scanning pattern (e.g., z-major order), and each element $s_i$ denotes a voxel located at slot index $i$. The index $i$ reflects the voxel's relative spatial position within the window, enabling the model to capture local geometric structure through consistent positional alignment.

Following the design of sparse convolutional networks~\cite{liu2015sparse}, we apply a sliding-window traversal over the 3D volume to determine where attention computation should occur. As illustrated in SWA U-Net of Fig.~\ref{fig:pipeline}, the window moves across the padded volume $\mathcal{V}$ with stride $S$. At each location, attention is computed only if the window contains at least one active (non-empty) voxel. This conditional activation reduces unnecessary computation and mirrors the sparsity-aware behavior of sparse convolutions. 

Compared to standard U-Nets that rely on pooling and interpolation, our SWA U-Net performs attention-based computations within these local windows. In the encoder, attention compresses input features by aggregating them over valid regions. In the decoder, attention propagates information into adjacent empty voxels. These positions are initialized with shared learnable embeddings, allowing the model to fill geometric gaps and densify the volume. This hierarchical transformation—from sparse to compact, and from sparse to dense—enables the network to complete large-scale semantic scenes.

\subsection{Intra-Window Spatial Embedding}

\begin{figure}[h]
	\centering
	\includegraphics[width=1\linewidth]{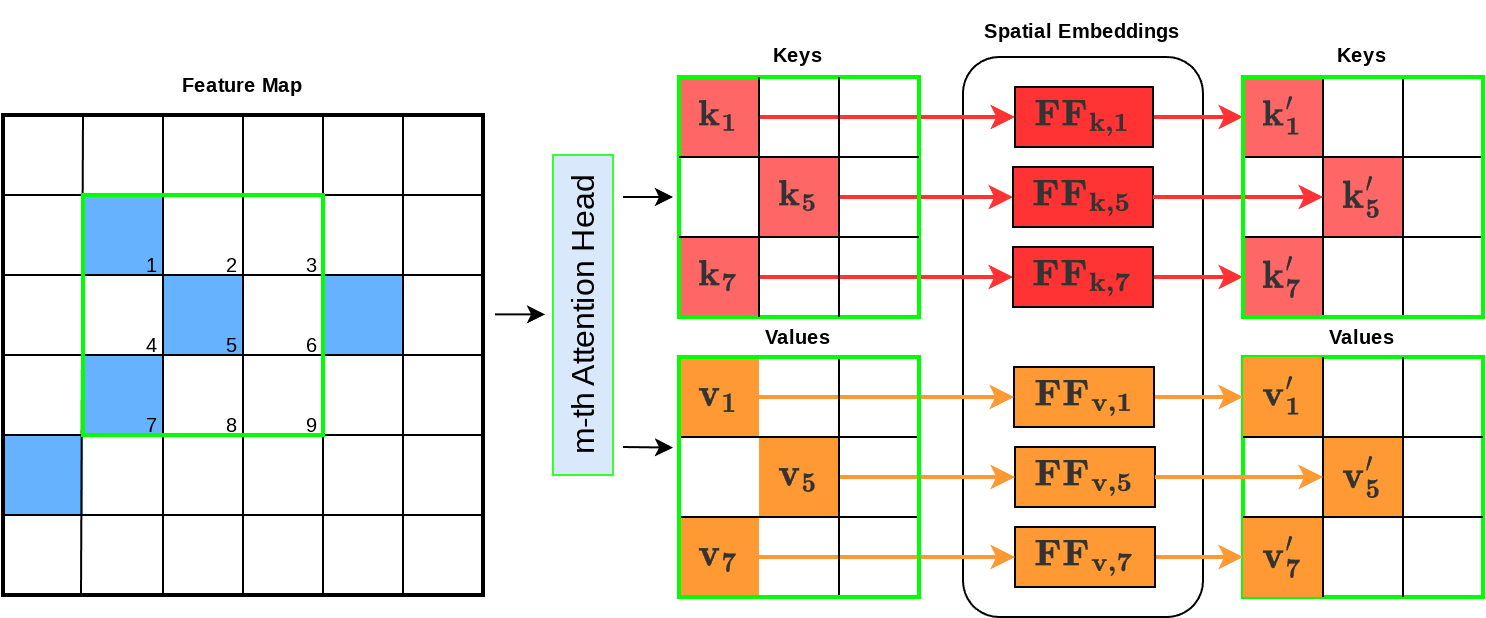}
	\caption{Illustration of the spatial embedding module. All valid voxels (highlighted in blue) are provided to each attention head with identical input features. Taking the $m$-th head as an example, spatial embedding is applied to each slot index through a dedicated feedforward layer that is shared across all heads, enabling the attention mechanism to integrate local spatial context in a position-dependent, but head-independent manner.
}
	\label{fig:spatial_embedding}
\end{figure}

To incorporate local spatial context into attention computation, we propose a spatial embedding mechanism, which is applied in addition to the standard multi-head attention, as shown in Fig.~\ref{fig:spatial_embedding}. Let $F \in \mathbb{R}^{L \times d}$ denote the input features of $L$ voxels in a window, each represented by a $d$-dimensional vector. For each attention head $m \in \{1, ..., M\}$, we compute head-specific key and value features using a dedicated feedforward layer followed by layer normalization:

\begin{equation}
K^{[m]} = \text{LN}(\text{FF}_k^{[m]}(F)), \quad V^{[m]} = \text{LN}(\text{FF}_v^{[m]}(F))
\end{equation}

Here, $K^{[m]} = [k_1^{[m]}, ..., k_L^{[m]}]$ and $V^{[m]} = [v_1^{[m]}, ..., v_L^{[m]}]$ denote the key and value vectors across all positions for head $m$. Each projection is shared across positions but varies with the attention head, enabling diverse representations.

To introduce spatial awareness, we apply a position-aware modulation to the key and value features. For each slot $i \in \{1, ..., L\}$, a dedicated feedforward layer $\text{FF}_i$ is applied to all heads at that position:

\begin{equation}
k_i'^{[m]} = \text{FF}_{k,i}(k_i^{[m]}), \quad v_i'^{[m]} = \text{FF}_{v,i}(v_i^{[m]})
\end{equation}

This position-aware modulation allows the attention mechanism to encode relative spatial position directly into the key and value features.

\subsection{Center Query}
To construct the query vector for attention, we adopt a center query strategy. Each attention window selects a center position $c$, which determines the query location for that window. If the center voxel is active, the query vector $q_c$ is computed from its input feature. Otherwise, a synthetic query is generated from the voxel’s global grid index using a separate MLP. This design ensures that a valid query vector is always available, even in sparsely populated regions. Formally, the query vector is defined as:
\begin{equation}
q_c =
\begin{cases}
\mathrm{FF}_q(f_c), & \text{if the}~f_c~\text{is non-empty} \\
\mathrm{FF}_q(p_c), & \text{otherwise}
\end{cases}
\end{equation}
where $f_c \in \mathbb{R}^d$ is the feature at voxel $c$, and $p_c \in \mathbb{Z}^3$ denotes its global grid index.

\begin{figure}[h]
	\centering
	\includegraphics[width=1\linewidth]{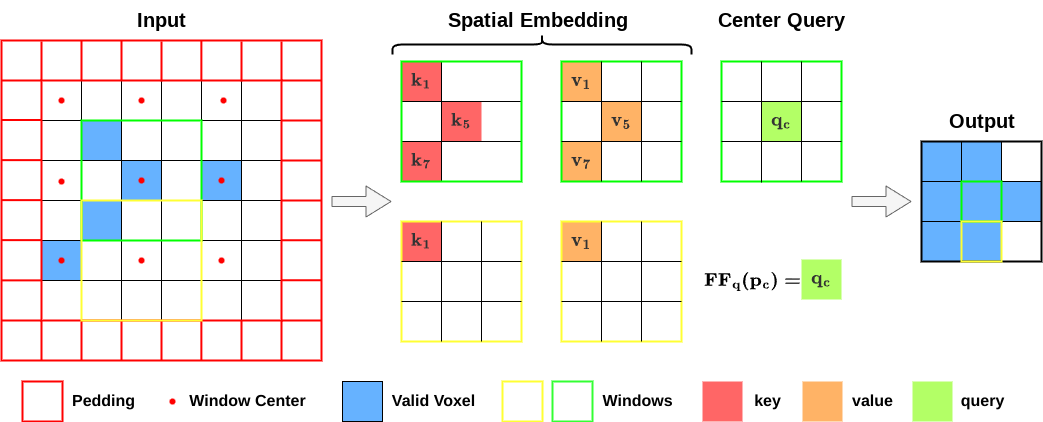}
	\caption{Overview of the intra- and inter-window attention pipeline. Windows traverse the voxel volume, and attention is computed only when active voxels are present inside the window. Keys and values are extracted from active elements with spatial embedding. The center query is generated either from the center voxel feature if present, or from its global grid index embedding if absent, and attends to all valid positions in the window.}
	\label{fig:center_query}
\end{figure}

Unlike spatial embedding, which encodes all valid voxels individually, the center query $q_c$ captures the contextual intent of the entire window and serves as a consistent anchor point across windows. When the center voxel is inactive, the center query still incorporates global context into the attention computation, allowing the model to integrate both global positioning and local spatial structure. The contrast between spatial embedding and center query, as well as the overall intra- and inter-window attention computation flow, is illustrated in Fig.~\ref{fig:center_query}.

\subsection{Attention Computation}
Our attention computation is built upon the combination of spatial embeddings and the center query strategy described above. Given a set of key vectors $\{k_i\}_{i=1}^L$ extracted from neighboring voxels within the window, we compute the attention energy at each position $i$ as the element-wise dot product between the center query $q_c$ and the corresponding key $k_i$:
\begin{equation}
e_i = q_c^\top k_i = \sum_{k=1}^d q_{c,k} \cdot k_{i,k}
\end{equation}

The energy values $\{e_i\}$ are then passed through a softmax operation to obtain normalized attention weights $\{\alpha_i\}$. To further enhance spatial adaptability, we introduce per-head modulation weights $\gamma^{[m]} \in \mathbb{R}^L$, where each element $\gamma_i^{[m]}$ reweights the attention score at position $i$ for attention head $m$:
\begin{equation}
\tilde{\alpha}_i^{[m]} = \gamma_i^{[m]} \cdot \alpha_i^{[m]}
\end{equation}

These spatial weights act analogously to learnable convolutional kernels, enabling each attention head to emphasize or suppress specific positions within the window.

\section{Evaluation}
\label{sec:Experiment}
\definecolor{carColor}{RGB}{100,150,245}
\definecolor{bicycleColor}{RGB}{100,230,245}
\definecolor{motorcycleColor}{RGB}{30,60,150}
\definecolor{truckColor}{RGB}{80,30,180}
\definecolor{othervehicleColor}{RGB}{0,0,255}
\definecolor{personColor}{RGB}{255,70,70}
\definecolor{bicyclistColor}{RGB}{255,40,200}
\definecolor{motorcyclistColor}{RGB}{150,30,90}
\definecolor{roadColor}{RGB}{255,0,255}
\definecolor{parkingColor}{RGB}{255,150,255}
\definecolor{sidewalkColor}{RGB}{75,0,75}
\definecolor{othergroundColor}{RGB}{175,0,75}
\definecolor{buildingColor}{RGB}{255,200,0}
\definecolor{fenceColor}{RGB}{255,120,50}
\definecolor{vegetationColor}{RGB}{0,175,0}
\definecolor{trunkColor}{RGB}{135,60,0}
\definecolor{terrainColor}{RGB}{150,240,80}
\definecolor{poleColor}{RGB}{255,240,150}
\definecolor{trafficsignColor}{RGB}{255,0,0}

\begin{table*}[t]
\caption{Quantitative results on SemanticKITTI hidden test set using LiDAR inputs.}
\vspace{-.5em}
\label{tab:evaluation}
\centering
\renewcommand{\arraystretch}{1.1} %<- modify value to suit your needs
\linespread{0.95}\selectfont
\setlength{\tabcolsep}{1.3pt}
\sisetup{separate-uncertainty}
\begin{threeparttable}
\begin{tabular}{c|c|ccccccccccccccccccc|c}%L{2.3cm}
  \toprule
  & SC & \multicolumn{20}{c}{SSC}\\
   Method & IoU & 
   \rotatebox{90}{\textbf{car} \tiny{(3.92$\%$)}} &
	\rotatebox{90}{\textbf{bicycle} \tiny{(0.03$\%$)}} &
	\rotatebox{90}{\textbf{motorcycle} \tiny{(0.03$\%$)}} &
	\rotatebox{90}{\textbf{truck} \tiny{(0.16$\%$)}} &
	\rotatebox{90}{\textbf{other-vehicle} \tiny{(0.20$\%$)}} &
	\rotatebox{90}{\textbf{person} \tiny{(0.07$\%$)}} &
	\rotatebox{90}{\textbf{bicyclist} \tiny{(0.07$\%$)}} &
	\rotatebox{90}{\textbf{motorcyclist} \tiny{(0.05$\%$)}} &
	\rotatebox{90}{\textbf{road} \tiny{(15.30$\%$)}} &
	\rotatebox{90}{\textbf{parking} \tiny{(1.12$\%$)}} &
	\rotatebox{90}{\textbf{sidewalk} \tiny{(11.13$\%$)}} &
	\rotatebox{90}{\textbf{other-ground} \tiny{(0.56$\%$)}} &
	\rotatebox{90}{\textbf{building} \tiny{(14.10$\%$)}} &
	\rotatebox{90}{\textbf{fence} \tiny{(3.90$\%$)}} &
	\rotatebox{90}{\textbf{vegetation} \tiny{(39.30$\%$)}} &
	\rotatebox{90}{\textbf{trunk} \tiny{(0.51$\%$)}} &
	\rotatebox{90}{\textbf{terrain} \tiny{(9.17$\%$)}} &
	\rotatebox{90}{\textbf{pole} \tiny{(0.29$\%$)}} &
	\rotatebox{90}{\textbf{traffic-sign} \tiny{(0.08$\%$)}} &
mIoU\\
  \midrule
   LMSCNet~\cite{roldao2020lmscnet}  & 
   56.7 &
   30.9 & 0.0 & 0.0 & 1.5 & 0.8 & 
   0.0 & 0.0 & 0.0 & 64.8 & 29.0 &
   34.7 & 4.6 & 38.1 & 21.3 & 41.3 &
   19.9 & 32.1 & 15.0 & 0.8 & 
   17.6 \\
   
   Local-DIFs~\cite{rist2021semantic}  & 
   57.7 &
   34.8 & 3.6 & 2.4 & 4.4 & 4.8 & 
   2.5 & 1.1 & 0.0 & 67.9 & \textbf{40.1} &
   \underline{42.9} & 11.4 & 40.4 & 29.0 & 42.2 &
   26.5 & 39.1 & 21.3 & 17.5 & 
   22.7 \\
   
   SSA-SC~\cite{yang2021semantic}  & 
   \textbf{58.8} &
   \textbf{36.5} & 13.9 & 4.6 & 5.7 & 7.4 & 
   4.4 & 2.6 & 0.7 & \textbf{72.2} & 37.4 &
   \textbf{43.7} & 10.9 & \underline{43.6} & 30.7 & \underline{43.5} &
   25.6 & \underline{41.8} & 14.5 & 6.9 & 
   23.5 \\
   
   JS3C-Net~\cite{yan2021sparse}  & 
   56.6 &
   33.3 & 14.4 & 8.8 & \underline{7.2} & 12.7 & 
   \underline{8.0} & \underline{5.1} & 0.4 & 64.7 & 34.9 &
   39.9 & \underline{14.1} & 39.4 & 30.4 & 43.1 &
   19.6 & 40.5 & 18.9 & 15.9 & 
   23.8 \\
   
   S3CNet~\cite{cheng2021s3cnet}  &
   45.6 &
   31.2 & \textbf{41.5} & \textbf{45.0} & 6.7 & \underline{16.1} & 
   \textbf{45.9} & \textbf{35.8} & \textbf{16.0} & 42.0 & 17.0 &
   22.5 & 7.9 & \textbf{52.2} & \underline{31.3} & 39.5 &
   \textbf{34.0} & 21.2 & \textbf{31.0} & \underline{24.3} &
   \textbf{29.5} \\
   
   SWA-SOP(Ours)  &
   \underline{57.9} &
   \underline{35.4} & \underline{17.2} & \underline{15.5} & \textbf{8.9} & \textbf{18.2} & 
   5.8 & 3.7 & \underline{1.3} & \underline{68.4} & \underline{39.0} &
   41.6 & \textbf{20.0} & 41.3 & \textbf{34.5} & \textbf{44.5} &
   \underline{31.2} & \textbf{42.2} & \underline{26.4} & \textbf{24.7} &
   \underline{27.4} \\
   \bottomrule
\end{tabular}
\end{threeparttable}

\vspace*{1ex} 
\footnotesize{IoU focuses solely on the occupancy status, while mIoU evaluates individual semantic categories. \textbf{Best} and \underline{second best} results are highlighted.}
\vspace*{-2ex}

\end{table*}

\subsection{Experiment Setup}

\subsubsection{Dataset}
We conduct our experiments on the SemanticKITTI dataset~\cite{behley2019iccv}, a widely used real-world autonomous driving dataset derived from the KITTI Odometry Benchmark~\cite{geiger2012cvpr}. It provides dense point-wise semantic annotations for large-scale outdoor driving environments. For semantic occupancy prediction, consecutive LiDAR scans are aggregated and voxelized to form dense semantic volumes, where each voxel is assigned a semantic label.

The input space is defined as $\mathcal{V}_{\text{kitti}} = \{(x, y, z) \mid x \in [0, 51.2]\,\text{m},\; y \in [-25.6, +25.6]\,\text{m},\; z \in [-2.0, 4.4]\,\text{m} \}$ and discretized into a regular voxel grid of size $256 \times 256 \times 32$, where each voxel represents a volume of $(0.2\,\text{m})^3$. Each voxel is labeled with one of 21 categories, including 19 semantic classes, empty space, and unknown regions. SemanticKITTI also provides an unknown mask for regions never directly observed from any scan position. These voxels are considered invalid and are excluded from both training and evaluation. The dataset comprises 22 sequences in total. Following the standard split, sequences 00--10 (excluding 08) are used for training, sequence 08 for validation, and sequences 11--21 for testing. Semantic occupancy prediction is evaluated using two standard metrics. The voxel-wise IoU focuses solely on binary occupancy status, measuring whether each voxel is occupied or free. In contrast, the mean IoU (mIoU) reflects the overall semantic segmentation accuracy across all valid categories within occupied regions.

\subsubsection{Implementation Details}
All models are trained for 50 epochs using the cross-entropy loss. We adopt a polynomial learning rate decay schedule with an initial learning rate of 0.006 and a decay power of 0.9. Each attention layer employs 8 heads with a hidden feature dimension of 128. Attention is computed within local windows of size \(3 \times 3 \times 3\), using a stride of 2 and padding of 1. Sparse convolution operations are implemented using the SpConv library~\cite{spconv2022}. Experiments on the point cloud pipeline are conducted using 6 NVIDIA TITAN RTX GPUs, while the RGB-based pipeline is trained with 4 GPUs.

\subsection{Main Results for LiDAR Input}
In Tab.~\ref{tab:evaluation}, we present the results of SWASOP on the hidden test set of the SemanticKITTI benchmark, compared to other state-of-the-art approaches. SWASOP integrates our proposed spatial embeddings and center queries. Our method achieves the second best IoU, slightly behind SSA-SC. In terms of mIoU, SWASOP also ranks second, marginally lower than S3CNet, and consistently ranks among the top across most semantic categories, often achieving either the best or second-best class-wise performance. It should be noted that S3CNet benefits from additional 2D BEV features, which significantly enhance the recognition of foreground objects. However, these 2D projections do not bring notable improvements to 3D scene completion and result in a significantly lower IoU compared to other baselines, indicating weaker capability in holistic occupancy prediction and leading to relatively inferior overall performance. In contrast, SWASOP maintains a more balanced and spatially consistent reconstruction, achieving strong results when considering both mIoU and IoU jointly.

\begin{figure*}[t]
  \centering
  \begin{tikzpicture}
    \node[anchor=south west,inner sep=0] (image) at (0,0) {\includegraphics[width=\textwidth]{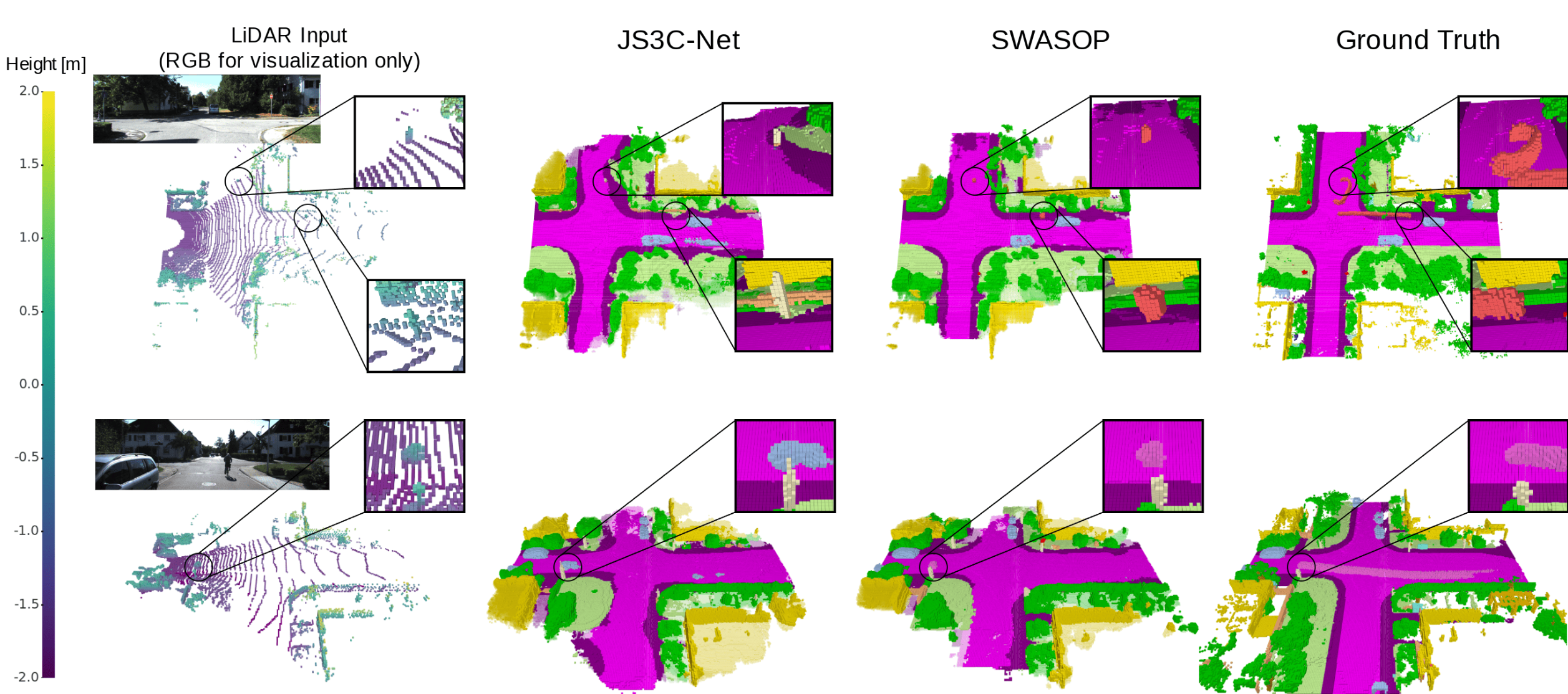}};
    \def\legendWidth{2.0}
    \def\legendHeight{0.2}
    \def\legendSpacing{0.15}
    \def\yOffset{-0.3}
    
   \draw[fill=carColor] ($ (image.south west) + (0.2, \yOffset)$) rectangle ++(0.2, 0.2) node[right, yshift=-0.1cm] {\small car};
	\draw[fill=bicycleColor] ($ (image.south west) + (1.2, \yOffset)$) rectangle ++(0.2, 0.2) node[right, yshift=-0.1cm] {\small bicycle};
    \draw[fill=motorcycleColor] ($ (image.south west) + (2.7, \yOffset)$) rectangle ++(0.2, 0.2) node[right, yshift=-0.1cm] {\small motorcycle};
    \draw[fill=truckColor] ($ (image.south west) + (4.7, \yOffset)$) rectangle ++(0.2, 0.2) node[right, yshift=-0.1cm] {\small truck};
    \draw[fill=othervehicleColor] ($ (image.south west) + (5.9, \yOffset)$) rectangle ++(0.2, 0.2) node[right, yshift=-0.1cm] {\small other-vehicle};
    \draw[fill=personColor] ($ (image.south west) + (8.1, \yOffset)$) rectangle ++(0.2, 0.2) node[right, yshift=-0.1cm] {\small person};
    \draw[fill=bicyclistColor] ($ (image.south west) + (9.5, \yOffset)$) rectangle ++(0.2, 0.2) node[right, yshift=-0.1cm] {\small bicyclist};
    \draw[fill=motorcyclistColor] ($ (image.south west) + (11.3, \yOffset)$) rectangle ++(0.2, 0.2) node[right, yshift=-0.1cm] {\small motorcyclist};
    \draw[fill=roadColor] ($ (image.south west) + (13.4, \yOffset)$) rectangle ++(0.2, 0.2) node[right, yshift=-0.1cm] {\small road};
    \draw[fill=parkingColor] ($ (image.south west) + (14.5, \yOffset)$) rectangle ++(0.2, 0.2) node[right, yshift=-0.1cm] {\small parking};
    \draw[fill=sidewalkColor] ($ (image.south west) + (16.0, \yOffset)$) rectangle ++(0.2, 0.2) node[right, yshift=-0.1cm] {\small sidewalk};
    
    \draw[fill=othergroundColor] ($ (image.south west) + (2.7, \yOffset-\legendHeight - \legendSpacing)$) rectangle ++(0.2, 0.2) node[right, yshift=-0.1cm] {\small other-ground};
    \draw[fill=buildingColor] ($ (image.south west) + (4.8, \yOffset-\legendHeight - \legendSpacing)$) rectangle ++(0.2, 0.2) node[right, yshift=-0.1cm] {\small building};
        \draw[fill=fenceColor] ($ (image.south west) + (6.3, \yOffset-\legendHeight - \legendSpacing)$) rectangle ++(0.2, 0.2) node[right, yshift=-0.1cm] {\small fence};
    \draw[fill=vegetationColor] ($ (image.south west) + (7.4, \yOffset-\legendHeight - \legendSpacing)$) rectangle ++(0.2, 0.2) node[right, yshift=-0.1cm] {\small vegetation};
    \draw[fill=trunkColor] ($ (image.south west) + (9.2, \yOffset-\legendHeight - \legendSpacing)$) rectangle ++(0.2, 0.2) node[right, yshift=-0.1cm] {\small trunk};
    \draw[fill=terrainColor] ($ (image.south west) + (10.4, \yOffset-\legendHeight - \legendSpacing)$) rectangle ++(0.2, 0.2) node[right, yshift=-0.1cm] {\small terrain};
    \draw[fill=poleColor] ($ (image.south west) + (11.7, \yOffset-\legendHeight - \legendSpacing)$) rectangle ++(0.2, 0.2) node[right, yshift=-0.1cm] {\small pole};
    \draw[fill=trafficsignColor] ($ (image.south west) + (12.8, \yOffset-\legendHeight - \legendSpacing)$) rectangle ++(0.2, 0.2) node[right, yshift=-0.1cm] {\small traffic-sign};
  \end{tikzpicture}
	
	\vspace*{-1ex}

  \caption{Qualitative results on the SemanticKITTI validation set. We show the voxelized LiDAR input (RGB image included for visualization only), the ground truth, and predictions from JS3C-Net for comparison. All 19 semantic classes are rendered without empty (void) regions. Predictions located in unknown areas are visualized with 20$\%$ opacity.}
  \label{fig:visualization}
  \vspace{-1.em}
\end{figure*}
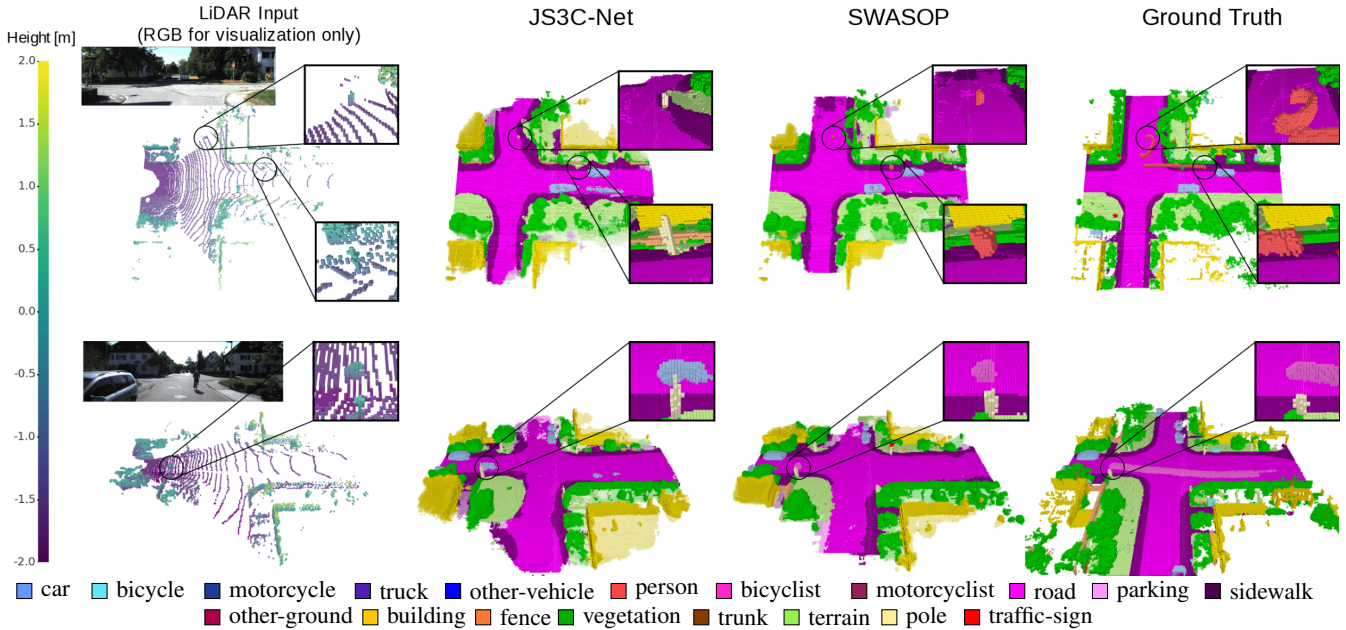

Fig.~\ref{fig:visualization} provides a qualitative comparison between SWASOP and JS3C-Net on selected scenes. In the first example, JS3C-Net incorrectly classifies a person as a pole, which also leads to a noticeable distortion in the predicted road geometry. In the second scene, a bicyclist is misclassified as a vehicle, resulting in shape and semantic inconsistency. It is worth noting that in the SemanticKITTI dataset, all dynamic objects are accumulated over time, which often leads to artifacts in the voxelized ground truth. Beyond improvements related to object-level recognition, SWASOP demonstrates overall superior performance in both the geometric prediction of foreground objects and the accurate reconstruction of large-scale background regions, highlighting its strength in producing consistent and detailed scene prediction.

\subsection{Extended Experiments for RGB Inputs}
To further demonstrate the generality of the proposed SWA module, we evaluate it as a plug-in component within an existing SOP pipeline based on RGB inputs. Specifically, we replace the original deformable self-attention module in VoxFormer with an adapted SWA U-Net, using the proposal queries generated in the first stage of VoxFormer as input voxels. This substitution is motivated by the observation that deformable attention relies on accurate depth-guided sampling and may suffer in far-range regions where the input becomes extremely sparse and the depth estimation is unreliable. In contrast, the SWA module leverages structured local attention to model geometric context within each window, offering enhanced robustness under such conditions. The results, reported in Tab.~\ref{tab:expand_experiment}, show that SWA consistently improves performance, indicating stronger scene prediction capabilities in sparse and uncertain areas. These findings highlight the module's effectiveness beyond LiDAR-based settings and its potential for integration into diverse perception architectures.

\begin{table}[h]
\caption{Quantitative results of our SWA module on the SemanticKITTI hidden test set using RGB inputs.}
\vspace{-.5em}
\label{tab:expand_experiment}
\centering
\setlength{\tabcolsep}{2pt}
\begin{tabular}{c|c|c|c|c}
	\toprule
	Method & MonoScene & VoxFormer-S & OccFormer & VoxFormer-S\\
	       & \cite{cao2022monoscene} & \cite{li2023voxformer} & \cite{zhang2023occformer} & + SWA \\
	\midrule
	mIoU & 11.08 & 12.20 & 12.32 & \textbf{13.19} \\
	IoU & 34.16 & 42.95 & 34.53 & \textbf{44.20} \\
	\midrule
	car & 18.8 & 20.8 & 21.6 & \textbf{23.3} \\
	bicycle & 0.5 & 1.0 & \textbf{1.5} & 0.9 \\
	motorcycle & 0.7 & 0.7 & \textbf{1.7} & 0.4 \\
	truck & 3.3 & 3.5 & 1.2 & \textbf{4.8} \\
	other-vehicle & \textbf{4.4} & 3.7 & 3.2 & 3.0 \\
	person & 1.0 & 1.4 & \textbf{2.2} & 1.1 \\
	bicyclist & 1.4 & \textbf{2.6} & 1.1 & 1.2 \\
	motorcyclist & \textbf{0.4} & 0.2 & 0.2 & 0.2 \\
	road & 54.7 & 53.9 & 55.9 & \textbf{56.0} \\
	parking & 24.8 & 21.1 & \textbf{31.5} & 23.6 \\
	sidewalk & 27.1 & 25.3 & \textbf{30.3} & 26.5 \\
	other-ground & 5.7 & 5.6 & 6.5 & \textbf{7.1} \\
	building & 14.4 & 19.8 & 15.7 & \textbf{21.5} \\
	fence & 11.1 & 11.1 & 11.9 & \textbf{13.3} \\
	vegetation & 14.9 & 22.4 & 16.8 & \textbf{25.0} \\
	trunk & 2.4 & 7.5 & 3.9 & \textbf{8.3} \\
	terrain & 19.5 & 21.3 & 21.3 & \textbf{22.2} \\
	pole & 3.3 & 5.1 & 3.8 & \textbf{5.9} \\
	traffic-sign & 2.1 &4.9 & 3.7 & \textbf{6.4} \\
	\bottomrule
\end{tabular}

\vspace*{1ex}
\footnotesize{IoU focuses solely on the occupancy status, while mIoU evaluates individual semantic categories. \textbf{Best} results are highlighted.}
\vspace*{-2ex}
\end{table}

\subsection{Ablation Studies}
We conduct an ablation study to investigate the individual contributions of the spatial embedding and center query components, with results summarized in Tab.~\ref{tab:ablation}. We use mIoU as the primary evaluation metric, as it most directly reflects semantic prediction quality in the SOP task. Three configurations (A, B, and C) are compared. Method A represents the full SWASOP architecture and achieves the best performance. Method B adopts the same spatial embedding for keys, values, and queries, without using the center query. The performance gap between A and B demonstrates the effectiveness of the center query, which provides a consistent and adaptive anchor for attention. When the center voxel is inactive, the query is instead generated from its global position embedding, enabling stable attention even in sparse regions. Method C disables the SWA module and directly uses a sparse convolutional U-Net. The lower performance confirms the superiority of our attention-based spatial modeling over traditional convolutional operations.

\begin{table}[h]
\caption{Ablation study of each sub-module on the SemanticKITTI validation set.}
\vspace{-.5em}
\label{tab:ablation}
\centering
\begin{tabular}{c|c|c|c}
	\toprule
	Method & A & B & C  \\
	\midrule
	Spatial Embedding   & \checkmark & \checkmark & -     \\
	Center Query      & \checkmark          & -          & -     \\
	\midrule
	mIoU                 & 27.91 & 27.07 & 24.46\\
	\bottomrule
\end{tabular}
\end{table}

%We will also measure the impact of our Augmented Label Rectification strategy on the performance of some state-of-the-art approaches.
%For the proposed ALR, we attempt to define reasonable maximum dimensions in voxels for each of the moving object classes, which will prevent the boxes surrounding them to grow beyond these bounds. The specific maximum bounds assigned to each class are enumerated in Tab. \ref{tab:ALR_max}.

\section{Conclusions}
In this work, we propose SWA-SOP, a novel semantic occupancy prediction framework. By adopting a sliding-window mechanism that selectively computes attention over valid regions, SWA-SOP significantly improves the efficiency of global attention-based models. Within each window, we leverage spatial embeddings and queries to perform localized attention, offering greater flexibility than traditional convolutional operations. Experiments on the SemanticKITTI benchmark demonstrate that our LiDAR-based method achieves state-of-the-art performance compared to existing approaches. Furthermore, we show that the proposed SWA module can be effectively integrated into image-based pipelines, leading to consistent performance gains. These results highlight the versatility of SWA and its potential for deployment across diverse sensor configurations in autonomous driving perception systems. In future work, we plan to further integrate SWA into recent advances in cross-modal fusion~\cite{wu2025llm, wuimgfu} and prompt-based representation learning~\cite{wu2025prompt}, enabling more scalable and flexible deployment in large-scale multimodal perception systems. Additionally, to support deployment on resource-constrained platforms, we aim to incorporate lightweight modeling techniques~\cite{li2024sglp, li2025frequency} to reduce computational cost without sacrificing performance.

%\IEEEtriggeratref{20}
%when using cite package:
\bibliographystyle{IEEEtran}
\bibliography{literature}

%\clearpage
%\newpage

\end{document}